\documentclass[10pt, a4paper]{article}
\usepackage{lrec}
\usepackage{multibib}
\newcites{languageresource}{Language Resources}
\usepackage{graphicx}
\usepackage{tabularx}
\usepackage{soul}
\usepackage{epstopdf}
\usepackage[latin1]{inputenc}
\usepackage{hyperref}
\usepackage{xstring}

\title{A Context-based Approach for Dialogue Act Recognition using Simple Recurrent Neural Networks}

\name{Chandrakant Bothe, Cornelius Weber, Sven Magg, and Stefan Wermter}

\address{Knowledge Technology, Department of Informatics, \\ University of Hamburg,\\
        Vogt-Koelln-Str. 30, 22527 Hamburg, Germany\\
        \url{http://www.informatik.uni-hamburg.de/WTM/} \\
        {\tt \{bothe,weber,magg,wermter\}@informatik.uni-hamburg.de}\\
        }

\abstract{
Dialogue act recognition is an important part of natural language understanding.
We investigate the way dialogue act corpora are annotated and the learning approaches used so far.
We find that the dialogue act is context-sensitive within the conversation for most of the classes.
Nevertheless, previous models of dialogue act classification work on the utterance-level and only very few consider context.
We propose a novel context-based learning method to classify dialogue acts using a character-level language model utterance representation, and we notice significant improvement.
We evaluate this method on the Switchboard Dialogue Act corpus, and our results show that the consideration of the preceding utterances as a a context of the current utterance improves dialogue act detection. \\ \newline  }

\begin{document}

\maketitleabstract

\section{Introduction}

In natural language processing research, the dialogue act (DA) concept plays an important role.
Its recognition, in most cases, is considered a lexical-based or syntax-based classification at utterance-level.
However, the discourse compositionality is context sensitive, meaning that the DA of an utterance can be elicited from the preceding utterances \cite{grosz1982discourse}. 
Hence, classifying only utterances is not enough because their DA class arises from their context. 
For example, the utterance containing only the lexical entry \textit{'yeah'} might appear in several DA classes such as \textit{Backchannel}, \textit{Yes-Answer}, etc.
For certain DA classes, the utterances are short, and most of them share similar lexical and syntactic cues \cite{Jurafsky1998}.

The aim of this article has two subgoals: first, we investigate the annotation process of DA corpora and review the modelling so far used for DA classification, and second, we present a novel model and compare its results with the state of the art.
We propose to use context-based learning for the identification of the DA classes. 
First, we show the results without context, i.e., classifying only utterances. Including context leads to 3\% higher accuracy. 
We use a simple recurrent neural network (RNN) for context learning of the discourse compositionality. 
We feed the preceding and current utterances to the RNN model to predict its DA class.
The main contributions of this work are as follows:

- We provide detailed insight on the annotation and modelling of dialogue act corpora. 
We suggest to model discourse within the context of a conversation.

- We propose a context-based learning approach for DA identification. In our approach, we represent utterances by a character-level language model trained on domain-independent data.

- We evaluate the model on the Switchboard Dialogue Act (SwDA\footnote{Available at \url{https://github.com/cgpotts/swda}}) corpus and show how using context affects the results. For the SwDA corpus, our model achieved an accuracy of 77.3\% compared to 73.9\% as state of the art, where the context-based learning is used for the DA classification \cite{kalchbrenner2013recurrent}.

- Benefits of using context arise from using only a few preceding utterances making the model suitable for dialogue system in real time, in contrast to feeding the whole conversation, which can achieve high accuracy, but includes future utterances \cite{ltAl2017EMNLP,kumar2017dasl}.

\begin{table*}[h]
\centering
\caption{Example of a labeled conversation (portions) from the Switchboard Dialogue Act corpus}
\label{sw_example}
\begin{tabular}{cll}
\hline
\textbf{Speaker} & \textbf{Dialogue Act} & \textbf{Utterance} \\ \hline
A                & Backchannel           & Uh-huh.\\
B                & Statement             & About twelve foot in diameter \\
B                & Abandoned             & and, there is a lot of pressure to get that much weight up in the air.\\
A                & Backchannel           & Oh, yeah.\\
B                & Abandoned             & So it's interesting, though.  \\
                 &                       & \ldots  \\
B                & Statement-opinion     & it's a very complex, uh, situation to go into space.\\
A                & Agree/Accept          & Oh, yeah,\\
                 &                       & \ldots  \\
A                & Yes-No Question       & You never think about that do you? \\
B                & Yes-Answer            & Yeah. \\
A                & Statement-opinion     & I would think it would be harder to get up than it would be \\
B                & Backchannel           & Yeah. \\ \hline
\end{tabular}
\end{table*}

\section{Related Work}

\subsection{Annotation of Dialogue Act Corpora}

\textbf{Annotation Process and Standards:} Research on dialogue acts became important with the commercial reality of spoken dialogue systems. 
There have been many taxonomies to it: speech act \cite{austin1962things} which was later modified into five classes (Assertive, Directive, Commissive, Expressive, Declarative) \cite{searle1979}, and the Dialogue Act Markup in Several Layers (DAMSL) tag set where each DA has a forward-looking function (such as Statement, Info-request, Thanking)
and a backward-looking function (such as Accept, Reject, Answer) \cite{allenCore1997}.
There are many such standard taxonomies and schemes to annotate conversational data, some of them follow the concept of discourse compositionality. 
These schemes are important for analysing dialogues or building a dialogue system \cite{skantze2007}.
However, there can never be a unique scheme that considers all aspects of dialogue.

\textbf{Corpus Insight:} We have investigated the annotation method for two corpora: Switchboard (SWBD) \cite{godfrey1992switchboard,shribergSwitch} and ICSI Meeting Recorder Dialogue Act (MRDA) \cite{shriberg2004icsi}. They are annotated with the DAMSL tag set. 
The annotation includes not only the utterance-level but also the segmented-utterance labelling.
The DAMSL tag set provides very fine-grained and detailed DA classes and follows the discourse compositionality.
For example, the SWBD-DAMSL is the variant of DAMSL specific to the Switchboard corpus.
It distinguishes \textit{wh-questions (qw)}, \textit{yes-no questions (qy)}, \textit{open-ended (qo)}, and \textit{or-questions (qr)} classes, not just because these questions are syntactically distinct, but also because they have different forward functions \cite{danjur1997swbddamsl}.
A \textit{yes-no question} is more likely to get a \textit{"yes"} answer than a \textit{wh-question}.
This also gives an intuition that the answers follow the syntactic formulation of question which provides a context.
For example \textit{qy} is used for a question that from a discourse perspective expects a \textit{Yes} or \textit{No} answer.

\textbf{Nature of Discourse in Conversation:} The dialogue act is a context-based discourse concept that means the DA class of a current utterance can be derived from its preceding utterance. 
We will elaborate this argument with an example given in Table \ref{sw_example}.
Speaker \textit{A} utters \textit{'Oh, yeah.'} twice in the first portion, and each time it is labelled with two different DA labels. 
This is simply due to the context of the previously conversed utterances.
If we see the last four utterances of the example, when speaker \textit{A} utters the \textit{'Yes-No Question'} DA, speaker \textit{B} answers with \textit{'yeah'} which is labelled as \textit{'Yes-Answer'} DA. 
However, after the \textit{'Statement-opinion'} from the same speaker, the same utterance \textit{'yeah'} is labelled as \textit{'Backchannel'} and not \textit{'Yes-Answer'}. 
This gives evidence that when we process the text of a conversation, we can see the context of a current utterance in the preceding utterances.

\textbf{Prosodic Cues for DA Recognition:} It has also been noted that prosodic knowledge plays a major role in DA identification for certain DA types \cite{Jurafsky1998,stolcke2000dialogue}.
The main reason is that the acoustic signal of the same utterance can be very different in a different DA class. 
This indicates that if one wants to classify DA classes only from the text, the context must be an important aspect to consider: simply classifying single utterances might not be enough, but considering the preceding utterances as a context is important.

\subsection{Modelling Approaches}

\textbf{Lexical, Prosodic, and Syntactic Cues:}
Many studies have been carried out to find out the lexical, prosodic and syntactic cues \cite{stolcke2000dialogue,surendran2006dialog,o2012multi,yang2014semi}. 
For the SwDA corpus, the state-of-the-art baseline result was 71\% for more than a decade using a standard Hidden Markov Model (HMM) with language features such as words and n-grams \cite{stolcke2000dialogue}.
The inter-annotator agreement accuracy for the same corpus is 84\%, and in this particular case, we are still far from achieving human accuracy.
However, words like \textit{'yeah'} appear in many classes such as \textit{backchannel}, \textit{yes-answer}, \textit{agree/accept} etc.
Here, the prosodic cues play a very important role in identifying the DA classes, as the same utterance can acoustically differ a lot which helps to distinguish the specific DA class \cite{shriberg1998can}. 
There are several approaches like traditional Naive Bayes and HMM models, which use minimal information and certainly ignore the dependency of the context within the communication \cite{grau2004dialogue,tavafi2013dialogue}. They achieved 66\% and 74.32\% respectively on the SwDA test set. 


\textbf{Utterance-level Classification:}
Perhaps most research in modelling dialogue act identification is conducted at utterance-level \cite{stolcke2000dialogue,grau2004dialogue,tavafi2013dialogue,ji2016latent,khanpour2016dialogue,lee2016sequential}. 
The emergence of deep learning also gave a big push to DA classification.
In a natural language conversation, most utterances are very short; hence it is also referred to as short text classification. 
Lee and Dernoncourt (2016) achieved 73.1\% accuracy on the SwDA corpus by using advanced deep learning frameworks such as RNNs and convolutional neural networks (CNN) with word-level feature embeddings.

\textbf{A Novel Approach: Context-based Learning:}
Classifying the DA classes at single utterance-level might fail when it comes to DA classes where the utterances share similar lexical and syntactic cues (words and phrases) like the \textit{backchannel}, \textit{yes-answer} and \textit{accept/agree} classes.
Some researchers proposed an utterance-dependent learning approach \cite{kalchbrenner2013recurrent,ji2016latent,kumar2017dasl,tzh2017EMNLP,ltAl2017EMNLP,ortega2017neural,Meng2017}. 
Kalchbrenner and Blunsom (2013) and Ortega and Vu (2017) have proposed context-based learning, where they represent the utterance as a compressed vector of the word embeddings using CNNs and use these utterance representations to model discourse within a conversation using RNNs. 
In their architecture, they also give importance to turn-taking by providing the speaker identity but do not analyse their model in this regard.
This approach achieves about 73.9\% accuracy on the SwDA corpus.
In another line of research \cite{ji2016latent,kumar2017dasl}, authors claim that their models take care of the dependency of the utterances within a conversation.
Ji et al. (2016) use discourse annotation for the word-level language modelling on the SwDA corpus and also highlight a limitation that this approach is not scalable to large data.

In other approaches a hierarchical convolutional and recurrent neural encoder model are used to learn utterance representation by feeding a whole conversation \cite{kumar2017dasl,ltAl2017EMNLP}.
The utterance representations are further used to classify DA classes using the conditional random field (CRF) as a linear classifier.
The model can see the past and future utterances at the same time within a conversation, which limits usage in a dialogue system where one can only perceive the preceding utterance as a context but does not know the upcoming utterances.
Hence, we use a context-based learning approach and regard the 73.9\% accuracy \cite{kalchbrenner2013recurrent} on the SwDA corpus as a current state of the art for this task.

\begin{figure}[t]
\begin{center}
\includegraphics[scale=0.85]{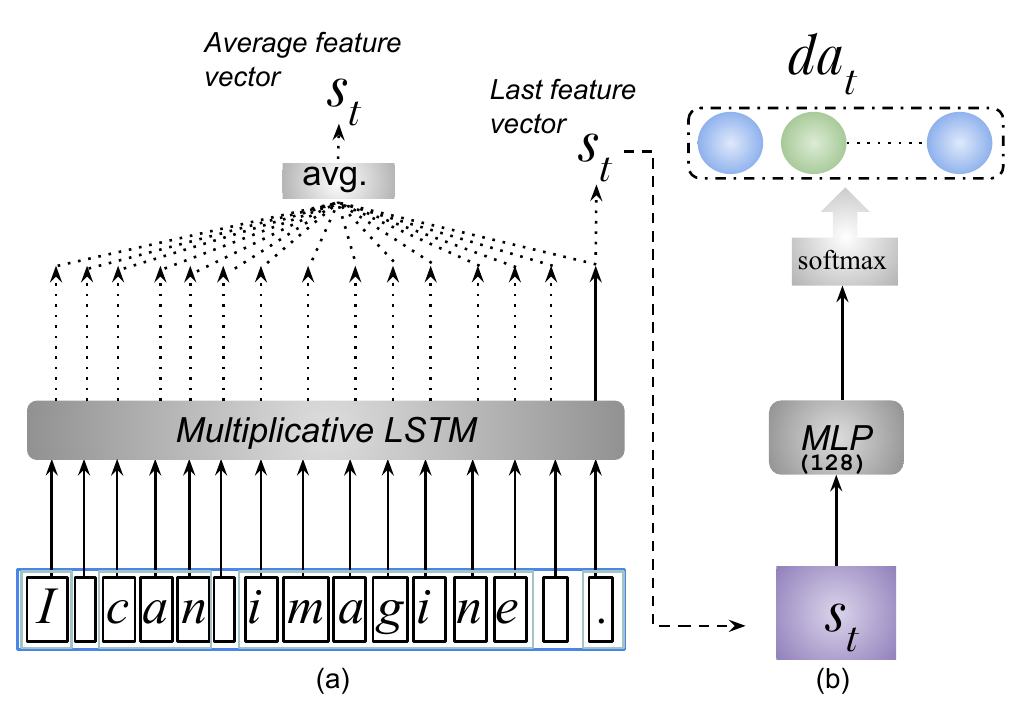} 
\caption{(a) Multiplicative LSTM (mLSTM) character-level language model to produce the sentence representation $s_{t}$.
The character-level language model is pre-trained and produces the feature (hidden unit states of mLSTM at the last character) or average (average of all hidden unit states of every character) vector representation of the given utterance. (b) Utterance-level classification using a simple MLP layer with a \textit{softmax} function (our baseline model).}
\label{sent_rep}
\end{center}
\end{figure}

\section{Our Approach}

Our approach takes care of discourse compositionality while recognising dialogue acts.
The DA class of the current utterance is predicted using the context of the preceding utterances.
We represent each utterance by the hidden state of the multiplicative recurrent neural network trained on domain-independent data using a character-level language model.
We use RNNs to feed the sequence of the utterances and eventually predict the DA class of the corresponding utterance.

\begin{figure}[t]
\begin{center}
\includegraphics[scale=0.8]{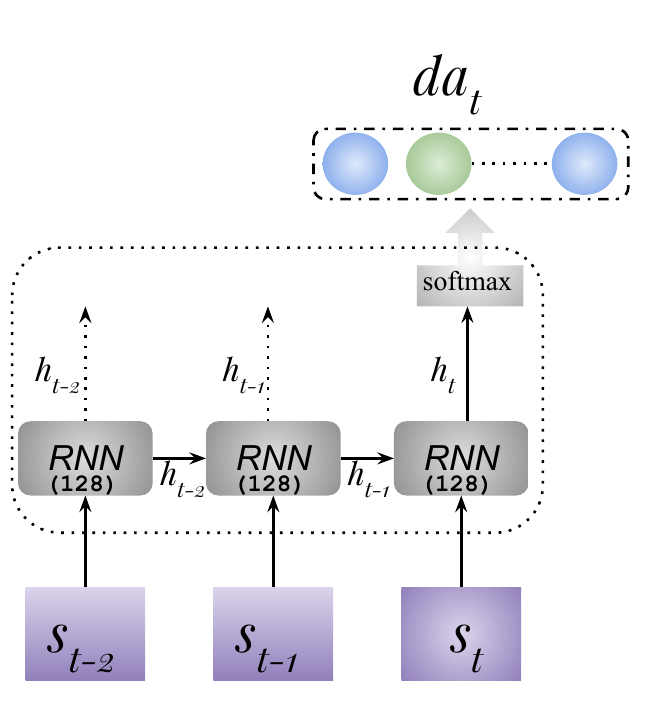} 
\caption{The RNN setup for learning the dialogue act recognition with
the previous sentences as context. 
$s_{t}$ is an utterance representation derived with a character-level language model and has a dialogue act label $da_{t}$. 
$s_{t-1}$ and  $s_{t-2}$ are the preceding utterances of  $s_{t}$. 
The RNN is trained to learn the recurrency through previous utterances $s_{t-1}$ and  $s_{t-2}$ derived as $h_{t-1}$ and $h_{t-2}$ as a context to recognize the dialogue act of current utterance $s_{t}$ which is represented by $h_{t}$ used to detect $da_{t}$.}
\label{rnn_setup}
\end{center}
\end{figure}

\subsection{Utterance Representation}

Character-level encoding allows processing words and whole sentences based on their smallest units and still capturing punctuation and permutation of words.
We represent a character-level utterance by encoding the whole sentence with a pre-trained character language model\footnote{\url{https://github.com/openai/generating-reviews-discovering-sentiment}}. 
This model consists of a single multiplicative long-short-term memory (mLSTM) network \cite{krause2016multiplicative} layer with 4,096 hidden units. 
The mLSTM is composed of an LSTM and a multiplicative RNN and considers each possible input in a recurrent transition function.
It is trained as a character language model on $\sim$80 million Amazon product reviews 
\cite{Radford2017LearningSentiment}.
We sequentially input the characters of an utterance to the mLSTM and take the hidden state values after the last character as shown in Figure \ref{sent_rep} (a). 

The hidden vector $s_{t}$ obtained after the last character is called the last feature vector, as it stores the information related to the character language model and the sentiment of the utterance. 
However, it was shown that the average vector over all characters in the utterance works better for emotion detection \cite{labothe2017WASSA2017}.
Hence, we extract the last feature vector and also the average feature vector representations for each utterance. 
We classify these representations with a multi-layer perceptron (MLP) as shown in Figure \ref{sent_rep} (b).
The results are shown in Table \ref{table:OldCharResults}. 
The standard deviation (SD) is computed over ten runs.
The average vector seems to carry more information related to the DA; hence we use it for future experiments. 
There is an advantage of using domain-independent data: it is rich regarding features being trained on big data, perhaps surpassing the limitation of scalability as mentioned in Ji et al. (2016). 

\subsection{Context Learning with RNNs}

We apply context learning with the help of RNNs. 
As shown in Figure \ref{rnn_setup}, the utterances with their character-level 
language model representation $s_t$ are fed to the RNN with the preceding utterances ($s_{t-1}$, $s_{t-2}$) being the context.
We use the RNN, which gets the input $s_t$, and stores the hidden vector $h_t$ at time $t$ \cite{elman1990finding}, which is calculated as:
\begin{equation}
h_{t}= f \left ( W_{h}*h_{t-1}+I*s_{t} + b \right )  
\end{equation} 
where $f()$ is a sigmoid function, $W_{h}$ and $I$ are recurrent and input weight matrices respectively and $b$ is a bias vector learned during training. 
$h_{t}$ is computed using the previous hidden vector $h_{t-1}$ which is computed in a same way for preceding utterance $s_{t-1}$.
The output $da_{t}$ is the dialogue act label of the current utterance $s_t$ calculated using $h_{t}$, as:
\begin{equation}
da_{t} = g \left ( W_{out}*h_{t} \right ) 
\end{equation}
where $W_{out}$ is the output weight matrix. 
The weight matrices are learned using back-propagation through time. 
The task is to classify several classes; hence we use a $softmax$ function $g()$ on the output. 
The input is the sequence of the current and preceding utterances, e.g., $s_t$, $s_{t-1}$, and $s_{t-2}$. 
We reset the RNN when it sees the current utterance $s_t$. 
We also give the information related to a speaker to let the network find the change in the speaker's turn.
The speaker id 'A' is represented by [1,0] and id 'B' by [0,1] and it is concatenated with the corresponding utterances $s_t$.

The Adam optimiser \cite{kingma2014adam} was used with a learning rate $1e-4$, which decays to zero during training, and clipping gradients at norm 1.  
Early stopping was used to avoid over-fitting of the network, 20\% of training samples were used for validation. 
In all learning cases, we minimise the categorical cross-entropy. 

\begin{table}[t]
\centering
\caption{Accuracy of the dialogue act identification using the character-level language model utterance representation for 42 classes using a single MLP layer with 64 neurons.}
\begin{tabular}{llll}
\hline
Model input                       &    Acc.(\%) & SD   \\

\hline
Last feature vector               &     71.48  & 0.28   \\
Average feature vector            & \bf 73.96  & 0.26  \\
Concatenated vector  \hspace{1cm} &     73.18  & 0.31   \\
\hline
\end{tabular}
\label{table:OldCharResults}
\end{table}

\subsection{Results}
We follow the same data split of 1115 training and 19 test conversations as in the baseline approach \cite{stolcke2000dialogue,kalchbrenner2013recurrent}.
Table \ref{table:charResults} shows the results of the proposed model with several setups, first without the context, then with one, two, and so on preceding utterances in the context.
We examined different values for the number of the hidden units of the RNN, empirically 64 was identified as best and used throughout the experiments. 
We also experimented with the various representations for the speaker id that is concatenated with the respective utterances but could find no differences. 
As a result, our proposed model uses minimal information for the context.
The performance increases from 74\% to about 77\% with context. 
We run each experiment for ten times and take the average.
The model shows robustness providing minimal variance, and using a minimum number of preceding utterances as a context can produce fair results.

\begin{table}[t]
\centering
\caption{Accuracy of the dialogue act identification with the context-learning approach. }
\begin{tabular}{llll}
\hline
Model setup                           &  Acc.(\%) & SD  \\
\hline
\textit{Baseline}                      &     \\
Most common class                      &  31.50   &  \\
\textit{Related previous work}         &          &   \\
Stolcke et al. (2000)                  &  71.00   &  \\
Kalchbrenner and Blunsom (2013)        &  73.90   &  \\

\textit{Our work}                      &          &     \\
Our baseline (without context)         &   73.96   & 0.26 \\
RNN (1 utt. in context w. SpeakerID)   &   76.48   & 0.33 \\
RNN (1 utt. in context)                &   76.57   & 0.28 \\
RNN (2 utts. in context) \hspace{1cm}  &   76.81   & 0.24 \\
RNN (3 utts. in context) \hspace{1cm}  & \bf 77.34 & 0.21 \\
RNN (4 utts. in context) \hspace{1cm}  &   77.28   & 0.22 \\ 
\hline
\end{tabular}
\label{table:charResults}
\end{table}

\section{Conclusion}

In this article, we detail the annotation and modelling of dialogue act corpora, and we find that there is a difference in the way DAs are annotated and the way they are modelled.
We argue to generalise the discourse modelling for conversation within the context of communication.
Hence, we propose to use the context-based learning approach for the DA identification task.
We used simple RNN to model the context of preceding utterances. 
We used the domain-independent pre-trained character language model to represent the utterances.
We evaluated the proposed model on the Switchboard Dialogue Act corpus and show the results with and without context. 
For this corpus, our model achieved an accuracy of 77.34\% with context compared to 73.96\% without context.
We also compare our model with Kalchbrenner and Blunsom (2013) who used the context-based learning approach achieving 73.9\%.
Our model uses minimal information, such as the context of a few preceding utterances which can be adapted to an online learning tool such as a spoken dialogue system where one can naturally see the preceding utterances but not the future ones.
This makes our model suitable for human-robot/computer interaction which can be easily plugged into any spoken dialogue system.

\section{Acknowledgements}

This project has received funding from the European Union's Horizon 2020 research and innovation programme under the Marie Sklodowska-Curie grant agreement number 642667 (SECURE).

\section{Bibliographical References}
\label{main:ref}

\bibliographystyle{lrec}
\bibliography{xample}


\section*{Appendix: Analysis of the state of the RNN}

\begin{figure*}[b]
\begin{center}
\includegraphics[width=17cm, height=11cm]{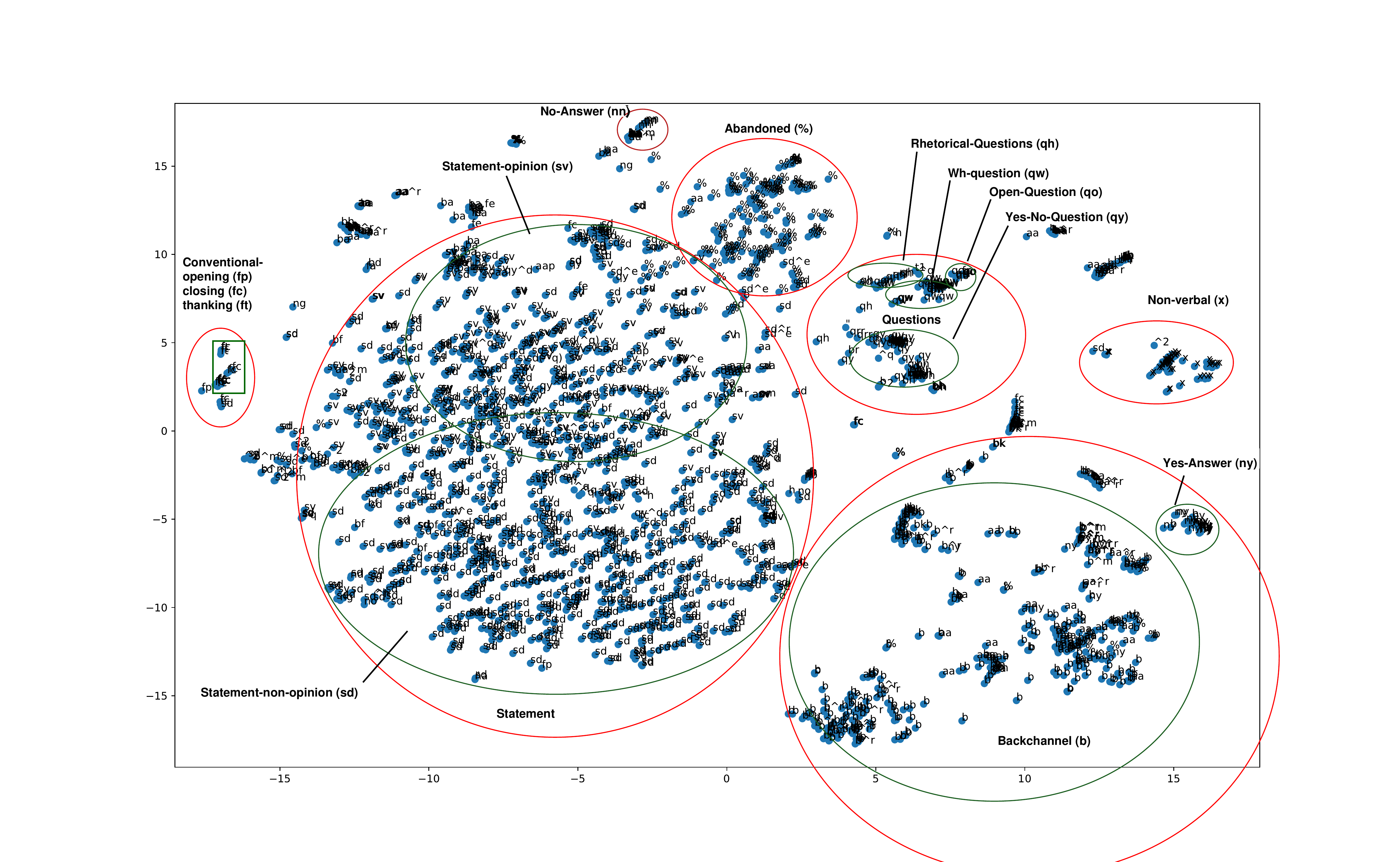} 
\caption{Clusters of all classes. Big clusters belong to the dominating \textit{Statement} classes, \textit{sv} and \textit{sd}.
The \textit{Question} classes, \textit{qy}, \textit{qw}, \textit{qh} and \textit{qo} are clustered within the big class. 
The classes \textit{Backchannel}, \textit{Yes-answers}, and \textit{Agree/Accept} share a lot of syntactic information hence they are clustered together, and our approach makes those classes separable within the cluster.}
\label{clusteringAll}
\end{center}
\end{figure*}

\begin{figure*}[b]
\begin{center}
\includegraphics[scale=0.9]{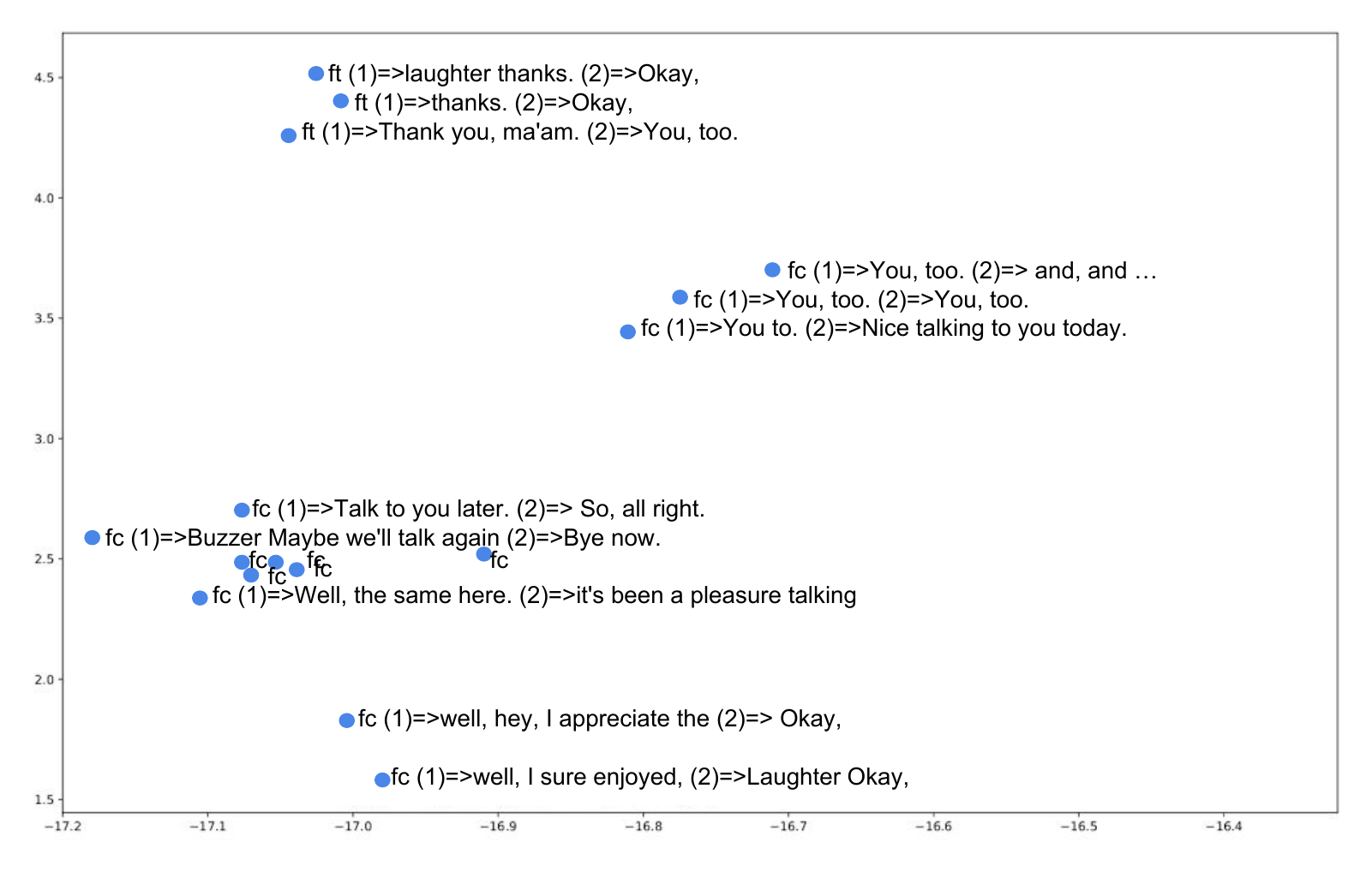} 
\caption{A blowup of the rectangle in Figure \ref{clusteringAll} from the \textit{Conventional closing (fc)} and \textit{thanking (ft) function} classes with their utterances. 
For readability, some utterances have been omitted and we show only the labels.
These are examples of the context-sensitive dialogues, where we can see one cluster of the \textit{ft} class and three groups of the \textit{fc} class. 
}
\label{fig_clusFfFt}
\end{center}
\end{figure*}

We also analyze the internal state $h_t$ of the RNNs for a two-utterance setup.
We plot them on a 2D graph with the t-SNE algorithm for the first 2,000 utterances of the SwDA test set.
Figure \ref{clusteringAll} shows the clusters of all the DA classes. 
The classes which do not share any information are grouped without any interference such as \textit{Non-verbal}, and \textit{Abandoned}.
Figure \ref{fig_clusFfFt} shows some particular classes with utterances in their vector spaces, the (1) current utterance and (2) a preceding utterance in the context.

\end{document}